\documentclass[journal,twoside,web]{ieeecolor2}
\usepackage{generic}
\usepackage{cite}
\usepackage{amsmath,amssymb,amsfonts}
\usepackage{algorithm2e}
\usepackage{graphicx}
\usepackage[hidelinks]{hyperref}
\usepackage{textcomp}
\usepackage{multirow}
\usepackage{orcidlink}
\usepackage{xcolor}
\definecolor{lemon}{RGB}{255,242,204}
\definecolor{lilac}{RGB}{231,205,255}
\definecolor{gainsboro}{RGB}{219,219,219}
\usepackage{subfigure}
\usepackage{pifont}
\def\BibTeX{{\rm B\kern-.05em{\sc i\kern-.025em b}\kern-.08em
    T\kern-.1667em\lower.7ex\hbox{E}\kern-.125emX}}
\begin{document}
\bstctlcite{IEEEexample:BSTcontrol}
\title{Personalizing Federated Instrument Segmentation with Visual Trait Priors \\
in Robotic Surgery}
\author{Jialang Xu\orcidlink{0000-0003-2324-7033}, Jiacheng Wang\orcidlink{0000-0003-2595-265X}, \IEEEmembership{Student Member, IEEE}, Lequan Yu\orcidlink{0000-0002-9315-6527}, \IEEEmembership{Member, IEEE}, Danail Stoyanov\orcidlink{0000-0002-0980-3227}, \IEEEmembership{Fellow, IEEE}, Yueming Jin\orcidlink{0000-0003-3775-3877}, \IEEEmembership{Member, IEEE}, and Evangelos B. Mazomenos\orcidlink{0000-0003-0357-5996}, \IEEEmembership{Member, IEEE}
\thanks{This work was supported in whole, or in part, by the Wellcome/EPSRC Centre for Interventional and Surgical Sciences (WEISS) [203145Z/16/Z and NS/A000050/1]; the EPSRC-funded UCL Centre for Doctoral Training in Intelligent, Integrated Imaging in Healthcare (i4health) [EP/S021930/1]; a UCL Research Excellence Scholarship; the Department of Science, Innovation and Technology (DSIT) and the Royal Academy of Engineering under the Chair in Emerging Technologies programme. For the purpose of open access, the authors have applied a CC BY public copyright licence to any author accepted manuscript version arising from this submission. \\ (\textit{Joint senior and corresponding authors: E. Mazomenos and Y. Jin})}
\thanks{J. Xu, D. Stoyanov and E. Mazomenos are with the Wellcome/EPSRC Centre for Interventional and Surgical Sciences and the Department of Medical Physics and Biomedical Engineering, University College London, London, UK (email: \{jialang.xu.22; danail.stoyanov; e.mazomenos\}@ucl.ac.uk).}
\thanks{J. Wang is with the Department of Computer Science, School of Informatics, Xiamen University, Xiamen, China (email: jiachengw@stu.xmu.edu.cn).}
\thanks{L. Yu is with the Department of Statistics and Actuarial Science, The University of Hong Kong, Hong Kong, China (email: lqyu@hku.hk).}
\thanks{Y. Jin is with the Department of Biomedical Engineering and the Department of Electrical and Computer Engineering, National University of Singapore, Singapore (email: ymjin@nus.edu.sg).}
}
\maketitle

\begin{abstract}
Personalized federated learning (PFL) for surgical instrument segmentation (SIS) is a promising approach. It enables multiple clinical sites to collaboratively train a series of models in privacy, with each model tailored to the individual distribution of each site. Existing PFL methods rarely consider the personalization of multi-headed self-attention, and do not account for appearance diversity and instrument shape similarity, both inherent in surgical scenes. We thus propose PFedSIS, a novel PFL method with visual trait priors for SIS, incorporating global-personalized disentanglement (GPD), appearance-regulation personalized enhancement (APE), and shape-similarity global enhancement (SGE), to boost SIS performance in each site. GPD represents the first attempt at head-wise assignment for multi-headed self-attention personalization. To preserve the unique appearance representation of each site and gradually leverage the inter-site difference, APE introduces appearance regulation and provides customized layer-wise aggregation solutions via hypernetworks for each site's personalized parameters. The mutual shape information of instruments is maintained and shared via SGE, which enhances the cross-style shape consistency on the image level and computes the shape-similarity contribution of each site on the prediction level for updating the global parameters. PFedSIS outperforms state-of-the-art methods with +1.51\% Dice, +2.11\% IoU, -2.79 ASSD, -15.55 HD95 performance gains. The corresponding code and models will be released at https://github.com/wzjialang/PFedSIS.
\end{abstract}


\section{Introduction}
\label{sec:introduction}
\IEEEPARstart{S}{urgical} instrument segmentation (SIS) is a critical task in robot-assisted surgery, enabling computer-aided navigation for improving outcomes~\cite{ross2021comparative}. Deep learning models have shown promise in SIS~\cite{sis1,gonzalez2020isinet,ternausnet}, but their effectiveness is influenced by the quantity and quality of available data~\cite{he2016deep}. Collaborative training, exploiting datasets from multiple sites, can boost the development and generalizability of SIS models. However, stringent privacy and confidentiality protocols make sharing surgical video recordings across sites impractical~\cite{fedmix,fedlc}. Federated learning (FL) has witnessed success in collaborative training for medical image segmentation tasks, where a global model is learned across multiple sites without data sharing, thereby alleviating privacy concerns~\cite{fedavg,sheller2019multi,fedbalanced,miao2023fedseg}.
FL methods though focus on a single global model, which is difficult to perform equally well for all sites, especially in surgical scenarios where different procedures, equipment, and patient characteristics introduce high heterogeneity. 

Personalized federated learning (PFL) overcomes this limitation by training multiple site-specific models rather than a single global model to better represent the data distribution of each site~\cite{pfl}. Typically PFL methods divide the architecture into global and personalized layers, with the latter composed of prediction heads~\cite{fedrep}, batch normalization~\cite{fedbn}, convolution channels~\cite{cd2pfedshen2022}, or query embeddings in self-attention~\cite{feddpwang2023}. Alternatives focus on the information of different sites, such as annotation inconsistency~\cite{feddpwang2023} and inter-site similarity~\cite{layerwisedma2022}. Though showing promising performance in applications on various medical modalities, such as prostate magnetic resonance imaging segmentation~\cite{iopfl}, lung computed tomography diagnosis~\cite{dou2021federated}, and low-count positron emission tomography denoising~\cite{zhou2023fedftn}, PFL for instrument segmentation from surgical videos remain under-explored. More importantly, we identify \textbf{two vital visual trait priors} in SIS: \emph{\textbf{instrument shape similarity}} and \emph{\textbf{surgical appearance discrepancy}}. Surgical instruments across different sites share a similar shape. For example, robotic instruments generally consist of three parts including shaft, wrist and jaws. Conversely, there exists a considerable gap in the appearance of surgical scenes. Leveraging both these priors in PFL for accurate SIS is important yet remains under-explored.

In this paper, we propose the first personalized federated learning method for surgical instrument segmentation that utilizes priors of instrument shape similarity and surgical appearance discrepancy to jointly train segmentation networks among multiple clinical sites, enhancing segmentation performance while ensuring data safety. The major contributions of this work are summarized as follows:
\begin{itemize}
    \item We propose \textbf{PFedSIS}, a novel PFL method for SIS, leveraging visual trait priors via three key components: (1) Global-personalized disentanglement \textbf{(GPD)}, decouples the parameter space into global and personalized by considering a fine-grained head-wise and channel-wise personalization for multi-headed self-attention (MSA) and convolution layers, respectively. This enables the model to represent a shared instrument structure via global parameters across sites, while styling the output with personalized parameters. To the best of our knowledge, our work is the original effort to personalize MSA at head-wise level for SIS; (2) Appearance-regulation personalized enhancement \textbf{(APE)} emphasizes the local site's distribution via appearance regulation and then learns and leverages hierarchical discrepancies across sites with hypernetwork-guided update; (3) Shape-similarity global enhancement \textbf{(SGE)} concentrates on mutual shape representation via cross-style shape consistency to enhance global parameters and aggregates them using a shape-similarity update.
    \item Experimental results on three publicly available benchmark datasets demonstrate our proposed PFedSIS surpasses state-of-the-art federated learning methods, achieving performance gains of 1.51\% to 2.28\% in Dice, 2.11\% to 3.08\% in IoU, -2.79 to -4.58 in ASSD, and -15.55 to -20.76 in HD95, while maintaining comparable computation complexity.
\end{itemize}

\section{Related Work}
\subsection{Federated Learning}
Federated learning (FL) enables multiple sites to jointly train a global model without sharing data, thereby alleviating data privacy concerns~\cite{fedavg}. Although FL has witnessed success in medical and surgical vision tasks via federated averaging~\cite{sheller2019multi}, balanced weights sharing strategies~\cite{fedbalanced}, dynamic weight averaging~\cite{shen2021multi}, style transfer~\cite{feddg}, and contrastive learning~\cite{miao2023fedseg}, these works focus on obtaining a single global model, which struggles to perform well across sites. This is prominent in surgical scenarios where different clinical settings, procedures, imaging systems, and patients introduce high heterogeneity in data distribution. In light of this limitation, personalized federated learning (PFL) is emerging as an appealing alternative. Unlike FL, which creates a single global model, PFL aims to produce individualized models that cater to the local distributions of each site’s data to enhance performance with privacy protection~\cite{pfl}. Many PFL approaches decouple the model architecture into global and personalized layers, where the global parameters are aggregated and shared via FedAvg~\cite{fedavg} at the server, while the personalized ones remain local. The global layer can be composed of prediction head layers~\cite{fedrep}, batch normalization layers~\cite{fedbn}, or parts of channels in convolutional layers~\cite{cd2pfedshen2022}. Some methods explore and utilize knowledge among different sites. For instance, FedDP~\cite{feddpwang2023} employs the inconsistencies in ground-truth annotations from various sites to calibrate local training. The pFedLA~\cite{layerwisedma2022} utilizes hypernetworks at the server, instead of distance metrics, to learn the similarities for all parameters between different sites. Unlike these methods, our novel approach focuses on separating global and personalized parameters following a multi-headed perspective. Moreover, it introduces shape similarity and appearance discrepancy priors for global and personalized parameters, respectively, thus providing site-independent shape and site-specific appearance information for each site.

\subsection{Surgical Instrument Segmentation}
Surgical instrument segmentation (SIS) aims to identify and segment surgical instruments in intra-operative scenes. Convolutional neural networks (CNNs) and U-Net architectures have been widely used for this task. For example, TernausNet~\cite{ternausnet} employs a U-Net-based network with a pre-trained VGG-11 or VGG-16 backbone~\cite{simonyan2014very} to predict surgical instruments using a pixel-based segmentation approach. ISINett~\cite{gonzalez2020isinet} introduces mask-based segmentation with Mask-RCNN~\cite{he2017mask} and a temporal consistency module for SIS. Recently, transformer-based methods have shown improved performance in SIS. STswinCL~\cite{jin2022explor} modifies the Swin Transformer~\cite{liu2021swin} with space-time shift and contrastive learning. MATIS~\cite{matis} incorporates long-term temporal information using MViT~\cite{fan2021multiscale} with Mask2Former~\cite{cheng2022masked}. However, the limited size of surgical datasets often results in models with insufficient generalization performance~\cite{yue2024surgicalsam}. More recent works~\cite{yue2024surgicalsam,zhou2023text} utilize vision foundation models like SAM~\cite{kirillov2023segment} and CLIP~\cite{clip} to enhance model generalization capabilities, through additional prompting or the incorporation of a language modality. Interestingly, previous work does not consider the potential benefits of leveraging visual trait priors from multiple datasets. We fill this gap and introduce PFedSIS, a novel PFL method based on shape similarity and appearance discrepancy priors for SIS, leveraging diverse datasets from multiple clinical sites to address data scarcity and privacy concerns.

\subsection{Hypernetworks}
Hypernetworks are deep neural networks tasked with generating the weights of a primary network, based on the input embedding they receive~\cite{shamsian2021personalized}. Hypernetworks are widely used in many fields, such as language modeling~\cite{mu2024learning}, meta-learning~\cite{beck2023hypernetworks}, continual learning~\cite{chandra2023continual}, few-shot learning~\cite{bertinetto2016learning}, personalized image generation~\cite{alaluf2022hyperstyle}, and multi-task learning~\cite{tay2020hypergrid}. In PFL, hypernetworks can generate weights to dynamically adapt model parameters based on site-specific embeddings, facilitating better personalization without compromising privacy. Shamsian et al.~\cite{shamsian2021personalized} is the first to introduce hypernetworks into FL. They proposed a hypernetwork with shared parameters for image classification tasks, where each local site possesses a unique embedding that serves as input, generating personalized model weights for different local sites. The pFedLA~\cite{layerwisedma2022} further refined hypernetworks for image classification to generate personalized weights for each model layer. In this paper, we propose APE, which includes hypernetworks that are updated based on the changes of personalized parameters, constrained by appearance regulation. APE generates the aggregation matrix tailored to each site's appearance distribution, demonstrating significant performance improvements for SIS.

\section{Methodology}
\subsection{Overall Pipeline}
\subsubsection{Problem Formulation}
We consider $M$ sites and their unique datasets $\{\mathcal{D}^m\}_{m=1}^M$, with each site connected to the server but only accessing its local dataset. The model at the $m$-th site has distinctive personalized parameters $\theta_P^m$ and shared global parameters $\theta_G$. The overall objective of PFL is:
\begin{equation}
\{\theta_G,\theta_P^1, \theta_P^2, \dots, \theta_P^M\}
=\mathop{\arg\min}\limits_{\{\theta_G,\theta_P^m\}_{m=1}^M} \sum_{m=1}^M k^m\mathcal{L}^m\left(\theta_G, \theta_P^m; \mathcal{D}^m\right)
\end{equation}
where balancing weight $k^m$ is typically set as $\frac{|\mathcal{D}^m|}{\sum\nolimits_{m=1}^M|D^m|}$,  $|\mathcal{D}^m|$ is the sample number of $m$-th local dataset, $\mathcal{L}^m(\theta_G, \theta_P^m; \mathcal{D}^m)$ denotes the loss function of site $m$.

\begin{figure*}[!tb]
    \centering
    \subfigure[]{
    \includegraphics[width=0.6\textwidth]{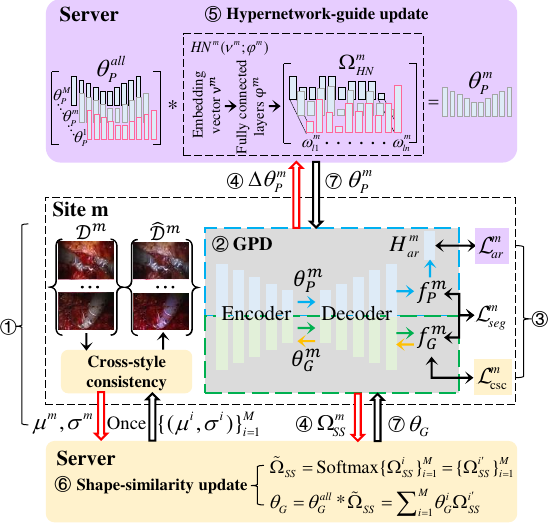}
    \label{fig1}
    }
    \subfigure[]{
    \includegraphics[width=0.29\textwidth]{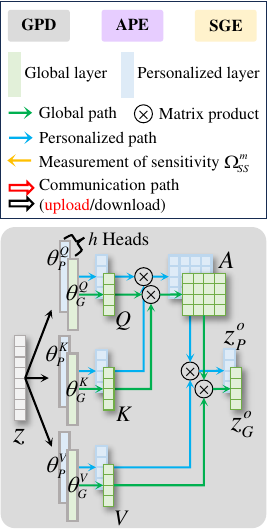}
    \label{fig_gpd}
    }
    \caption{Our proposed PFedSIS. (a) The overview of the PFedSIS architecture. The global-personalized disentanglement (GPD), appearance-regulation personalized enhancement (APE), and shape-similarity global enhancement (SGE) modules are highlighted with \colorbox{gainsboro}{{\strut gainsboro}}, \colorbox{lilac}{{\strut lilac}}, and \colorbox{lemon}{{\strut lemon}} backgrounds, respectively. In APE, considering site $m$'s hypernetwork $HN^m(\nu^m;\varphi^m)$ as an example, $\nu^m$ and $\varphi^m$ are updated based on the change of site $m$'s personalized parameters $\Delta \theta_P^m$. \ding{172}--\ding{178} represent the workflow steps of PFedSIS; (b) Illustrations of the multi-headed self-attention in GPD.}
\end{figure*}

\subsubsection{Overview of PFedSIS}
Fig.~\ref{fig1} illustrates the PFedSIS architecture, consisting of a transformer-based PVTv2~\cite{wang2021pvtv2} encoder and a FPN~\cite{fpn} convolutional decoder with an appearance regulation head $H_{ar}$. The proposed \textbf{GPD} decouples the encoder and decoder in personalized and global parameters. The \textbf{APE} comprises appearance regulation and hypernetwork-guided update, while the \textbf{SGE} involves cross-style consistency and shape-similarity update. Considering site $m$ from the total $M$ sites, the pseudocode of PFedSIS is presented in Algorithm~\ref{alg_pfedsis}, and its workflow is detailed as follows:
\begin{enumerate}
\renewcommand{\labelenumi}{\ding{\numexpr171+\value{enumi}\relax}}
    \item Each site uploads the mean $\mu$ and standard deviation $\sigma$ of its local dataset $\mathcal{D}$ to formulate a style-memory set $\{(\mu^i, \sigma^i)\}^M_{i=1}$, and site $m$ downloads it only once to generate a style-distorted dataset $\hat{\mathcal{D}}^m$ via cross-style consistency.
    \item Disentanglement of the model's architecture into personalized $\theta_P^m$ and global $\theta_G^m$ via GPD.
    \item Input selection in pairs of original images and corresponding style-distorted images, from $\mathcal{D}^m$ and $\hat{\mathcal{D}}^m$. Local training via cross-entropy segmentation loss $\mathcal{L}_{seg}^m$, cross-style shape consistency loss $\mathcal{L}_{csc}^m$, and appearance regulation loss $\mathcal{L}_{ar}^m$. We prioritize $\theta_{P}^m$ on the appearance information specific to site $m$, while $\theta_{G}^m$ focuses on acquiring shape-similarity features across sites.
    \item Site $m$ uploads the change of personalized parameters $\Delta\theta_P^m$ and sensitivity $\Omega_{SS}^m$ to the server.
    \item Server updates the layer-wise aggregation matrix $\Omega_{HM}^m$ of site $m$, with the corresponding hypernetwork $HN^m(\nu^m;\varphi^m)$ according to $\Delta\theta_P^m$, and updates personalized parameters $\theta_P^m$ based on $\Omega_{HM}^m$.
    \item Server computes the shape-similarity matrix $\tilde\Omega_{SS}$ based on the collection $\{\Omega_{SS}^{i}\}_{i=1}^M$ uploaded by all sites, and updates global parameters $\theta_G$ based on $\tilde\Omega_{SS}$.
    \item Site $m$ downloads $\{\theta_G, \theta_P^m\}$ and returns to step \ding{174} for the next local training.
\end{enumerate}
\RestyleAlgo{ruled}
\begin{algorithm}[!tbh]
\caption{Pseudocode of the proposed PFedSIS}\label{alg_pfedsis}
\KwIn{$M$ sites $\{C^m\}_{m=1}^M$ with their dataset $\left\{\mathcal{D}^m\right\}_{m=1}^M$, total communication rounds $T$, total local training iterations $L$, local learning rate $\eta$, initial model parameters of all $M$ sites $\{\theta_G^{m,(0)},\theta_P^{m,(0)}\}_{m=1}^M$, initial $M$ hypernetworks $\{HN^{m,(0)}(\nu^m;\varphi^m)\}_{m=1}^M$.}
\KwOut{Reasonable $M$ personalized models $\{\theta_G,\theta_P^m\}_{m=1}^M$.}
\tcc{\textbf{Server}}
\ding{172} Format the style-memory set $\{(\mu^i,\sigma^i)\}_{i=1}^M$\;
\For{communication round $t=0$ \KwTo $(T-1)$}{
        \For{each site $m \in \{C^m\}_{m=1}^M$ in parallel}{
        \tcc{\textbf{Site}}
        \ding{178} Site $m$ downloads $\theta_G^{(t)}$ and $\theta_P^{m,(t)}$ from server\;
        \ding{172} Generated $\hat{\mathcal{D}}^m$ based on $\mathcal{D}^m$ via Eq.~\eqref{csc_mix}\;
        \ding{173} Set $\theta_G^{m,(0)},\theta_P^{m,(0)} \gets \theta_G^{(t)},\theta_P^{m,(t)}$\;
        \For{local iteration $l=0$ \KwTo $(L-1)$}{
        \tcp{\ding{174} Local Training}
        Sample mini-batch $(I,\hat{I}) \subseteq (\mathcal{D}^m, \hat{\mathcal{D}}^m)$\;
        Update $\theta_P^{m,(l+1)}\gets\theta_P^{m,(l)}-\eta\nabla_{\theta_P^{m,(l)}}\left(\mathcal{L}_{seg}(I) + \mathcal{L}_{ar}(I)\right)$\;
        Update $\theta_G^{m,(l+1)}\gets\theta_G^{m,(l)}-\eta\nabla_{\theta_G^{m,(l)}}\left(\mathcal{L}_{seg}(I) + \mathcal{L}_{csc}(\hat{I})\right)$\;
        }
        Compute $\Delta \theta_P^m \gets \theta_P^{m,(L)}-\theta_P^{m,(0)}$\;
        Compute $\Omega_{SS}^m$ via Eq.~\eqref{ssm}\;
        \ding{175} Upload $\Delta \theta_P^{m}, \Omega_{SS}^m, \theta_G^{m,(L)}, \theta_P^{m,(L)}$ to server\;
        \tcc{\ding{176} \textbf{Server}}
        Set $\theta_{P}^{all,(t+1)} \gets \{\theta_{P}^{1,(L)}, \theta_{P}^{2,(L)}, \dots, \theta_{P}^{M,(L)}\}$\;
        Update $HN^{m,(t+1)}(\nu^m;\varphi^m)$ via Eqs.~\eqref{hn_update_1}\eqref{hn_update_2}\;
        $\Omega_{HM}^{m,(t+1)} \gets HN^{m,(t+1)}(\nu^m;\varphi^m)$\;
        Update $\theta_{P}^{m,(t+1)}$ via Eq.~\eqref{hpa_up}\;
        }
        \tcc{\ding{177} \textbf{Server}}
        Set $\theta_{G}^{all,(t+1)} \gets \{\theta_{G}^{1,(L)}, \theta_{G}^{2,(L)}, \dots, \theta_{G}^{M,(L)}\}$\;
        Compute $\tilde{\Omega}_{SS}^{(t+1)}$ via Eq.~\eqref{ssm_server}\;
        Update $\theta_G^{(t+1)}$ shared by all sites via Eq.~\eqref{gp_server};
        }
\end{algorithm}

\subsection{Global-Personalized Disentanglement (GPD)}
To attain global shape similarity among instruments while mitigating appearance heterogeneity across sites, we propose a GPD strategy, depicted in Fig.~\ref{fig1}. As multiple embedding heads capture different representation subspaces~\cite{msa}, GPD decouples half of the query, key, value embedding heads in the MSA, half of the convolution channels in the FPN decoder, and the entire appearance regulation head $H_{ar}$ as personalized parameters ($\theta_{P}$), and designating the remaining half as global parameters ($\theta_{G}$). As shown in Fig.~\ref{fig_gpd}, for MSA with $h$ heads, given $z\in\mathbb{R}^L$ as the input feature with length $L$, the query $Q_i$, key $K_i$, and value $V_i$ from $i$-th head are formulated as:
\begin{equation}
\begin{aligned}
head_i(z) &= \{Q_i, K_i, V_i\} \\
&=
\begin{cases}
\{z\theta_{P}^Q, z\theta_{P}^K, z\theta_{P}^V\}, & i \in \{1, \dots, \frac{h}{2}\} \\
\{z\theta_{G}^Q, z\theta_{G}^K, z\theta_{G}^V\}, & \text{others}
\end{cases}
\end{aligned}
\end{equation}
where $\theta_{P}^{Q/K/V}, \theta_{G}^{Q/K/V} \in \mathbb{R}^{L \times d_{head}}$ denote linear projections of personalized and global parameters for the query/key/value embedding heads, respectively. The dimension of each head $d_{head}$ is equal to $L/h$.

Then, the output $z_i^{o} \in \mathbb{R}^{d_{head}}$ of the $i$-th head can be computed by element-wise multiplication of the similarity matrix $A_i$ with $V_i$, where $A_i$ is obtained by comparing each pixel in query $Q_i$ to all the elements in $K_i$ in the long-range view: 
\begin{equation}
   z_i^{o} = A_i V_i = Softmax(\frac{Q_i K_i^T}{\sqrt{d_{head}}})V_i
\end{equation}

The outputs of MSA $z_P^{o}, z_G^{o}\in \mathbb{R}^{d_{head}\times \frac{h}{2}}$ are derived by concatenating the outputs of the first $\frac{h}{2}$ and the last $\frac{h}{2}$ heads, respectively:
\begin{equation}
\begin{aligned}
GDP_{MSA}(z) &= \{z_P^{o},z_G^{o}\} \\
&=\{\operatorname{Concat}(z_1^{o}, \dots, z_{\frac{h}{2}}^{o}), \operatorname{Concat}(z_{\frac{h}{2}+1}^{o}, \dots, z_{h}^{o})\}
\end{aligned}
\end{equation}

Finally, $z_P^{o}$ and $z_G^{o}$ are fed into the personalized and global channels of the FPN decoder to obtain personalized $f_P$ and global features $f_G$, respectively. 

\subsection{Appearance-regulation Personalized Enhancement (APE)}
\subsubsection{Appearance regulation}
The diversity in background, color, texture, camera settings, and anatomical locations in robot-assisted surgery leads to substantial variations in appearance across different sites. We thus develop a constraint in the personalized parameters $\theta_{P}$, to intensify attention on the distinctive appearance characteristics of the local dataset. Specifically, we use an appearance regulation head $H_{ar}$ after the decoder to perform image reconstruction from the personalized features $f_P$, defined as:
\begin{equation}
    \mathcal{L}_{ar}(I)=\|H_{ar}(f_P) - I\|_2^{2}
\end{equation}
where $I$ is the input images and $\|\cdot\|_2^{2}$ is squared L2 norm.

\subsubsection{Hypernetwork-guided update}
After applying appearance regulation to $\theta_{P}$, we train specialized hypernetworks at the server. Each site is assigned a hypernetwork to generate an aggregation matrix for customizing its $\theta_{P}$ to gradually leverage inter-site appearance discrepancy at layer level. As shown in Fig.~\ref{fig1}, taking  site $m$ as an example, the hypernetwork $HN^m(\nu^m;\varphi^m)$ consists of a learnable embedding vector $\nu^m$ as the descriptor of site $m$ and multiple fully connected layers $\varphi^m$, which are alternatively trained with the personalized parameters, to output a layer-wise aggregation matrix $\Omega_{HN}^m$ for personalized aggregation:
\begin{equation}
\label{am}
\Omega_{HN}^m = [\omega_{l1}^m, \omega_{l2}^m, \dots, \omega_{ln}^m]
\end{equation}
where $\omega_{ln}^m$= $[\omega_{ln,1}^m, \omega_{ln,2}^m, \dots, \omega_{ln,M}^m]^T$ is the aggregation vector of the $n$-th personalized layer in site $m$, and $\omega_{ln,M}^m$ is the aggregation weight for site $M$ in $n$-th personalized layer. Finally, the personalized parameters $\theta_{P}^{m}$ in site $m$ are updated at the server by weighted aggregation based on $\Omega_{HN}^m$:
\begin{equation}
\label{hpa_up}
\theta_{P}^{m} = \theta_{P}^{all} * \Omega_{HN}^m = \{\sum_{i=1}^M \theta_{P}^{l1,i}\omega_{l1,i}^m, \dots, \sum_{i=1}^M \theta_{P}^{ln,i}\omega_{ln,i}^m\}
\end{equation}
where $\theta_P^{all} = \{\theta_{P}^{i}\}_{i=1}^M$ is the set of personalized parameters from all $M$ sites, while $\theta_{P}^{i}=\{\theta_{P}^{l1,i}, \theta_{P}^{l2,i}, \dots, \theta_{P}^{ln,i}\}$ is the set of personalized parameters from all personalized layers from site $i$. Similar to~\cite{shamsian2021personalized}, we adopt a universal way to update the hypernetwork $HN^m(\nu^m;\varphi^m)$:
\begin{equation}
\label{hn_update_1}
\Delta \nu^m = (\nabla_{\nu^m} \theta_P^m)^T \Delta \theta_P^m =\left(\theta_{P}^{all} * \nabla_{\nu^m}\Omega_{HN}^m\right)^T
\Delta \theta_P^m
\end{equation}
\begin{equation}
\label{hn_update_2}
\Delta \varphi^m = (\nabla_{\varphi^m} \theta_P^m)^T \Delta \theta_P^m =\left(\theta_{P}^{all} * \nabla_{\varphi^m}\Omega_{HN}^m\right)^T
\Delta \theta_P^m
\end{equation}
where $\Delta \theta_P^m$ is the change of personalized parameters at site $m$ after local training. Note that the $\theta_P^m$ is partially updated by appearance regulation loss $\mathcal{L}_{ar}^m$, thus the update direction of $HN^m(\nu^m;\varphi^m)$ aligns with the optimization direction of the appearance reconstruction objective to consider appearance-related information.

\subsection{Shape-similarity Global Enhancement (SGE)}
\subsubsection{Cross-style shape consistency}
To maintain robust shape representation against varied surgical scenes for the global parameters $\theta_{G}$, we propose cross-style shape consistency, which makes $\theta_{G}$ segment the instrument under the style perturbation of multiple sites. More specifically, as style is linked to the mean and standard deviation~\cite{adain,mixstyle,xu2023regressing}, for site $m$, we first get the inter-site style statistics $(\beta_{cross}^m, \gamma_{cross}^m)$ by mixing the mean and standard deviation of all sites:
\begin{equation}
\beta_{cross}=\sum\nolimits_{i=1}^M \lambda_i^m \mu^i; \gamma_{cross}=\sum\nolimits_{i=1} \lambda_i^m \sigma^i
\end{equation}
where $\mu^i$ and $\sigma^i$ are the mean and standard deviation of the $i$-th site from the style-memory set $\{(\mu^i, \sigma^i)\}^M_{i=1}$, representing the characteristic of each site. $\{\lambda_i^m\}_{i=1}^M$ are random convex values sampled from $\operatorname{Beta}(0.1,0.1)$. Note that $\{(\mu^i, \sigma^i)\}^M_{i=1}$ are $2 \times M$ scalar values and recovery of images solely based on the mean and standard deviation is unfeasible~\cite{cross2023Chen}, ensuring privacy protection.

Then we apply $(\beta_{cross}^m, \gamma_{cross}^m)$ to distort the original input $I$ and adopt segmentation ground truth $Y$ of $I$ as the supervising signal for local training to enforce shape consistency on the style-distorted input  $\hat{I}$ at the prediction level. Formally, the cross-style shape consistency is defined as:
\begin{align}
\hat{I} =\gamma_{cross}^m \frac{I-\mu^{m}}{\sigma^m}+\beta_{cross}^m \label{csc_mix} \\
\mathcal{L}_{csc}(\hat{I})=\|\theta_G^m(\hat{I}) - Y\|_2^{2} \label{csc_loss}
\end{align}
where $I\in \mathcal{D}^m$, with $\mathcal{D}^m$ representing the dataset of site $m$, $\theta_G^m(\hat{I})$ is the segmentation prediction for $\hat{I}$ from $m$-th site global parameters $\theta_G^m$.

\subsubsection{Shape-similarity update}
To share and preserve shape knowledge among different sites, we access the sensitivity of shape-related predictions (i.e. segmentation predictions) to global parameter perturbations in the model, generating a shape-similarity matrix $\tilde\Omega_{SS}$ at the server to aggregate and update global parameters from all sites. Segmentation predictions contain crucial information about the shape of surgical instruments, which should be preserved during the aggregation of global parameters. Inspired by the neural plasticity mechanisms in the human brain that regulate memory retention and forgetting \cite{fusi2005cascade,s3rzhang2023}, we estimate how sensitive segmentation predictions are to changes in global parameters to guide the global parameter aggregation at the server. Specifically, for the $m$-th site with a total of $N$ images, it can be formulated as follows:
\begin{equation}
\label{ssm}
\Omega_{SS}^m =\frac{1}{N} \sum\nolimits_{i=1}^N \nabla_{\theta_G^m}\left\|\theta_G^m(I_i)\right\|_2^2
\end{equation}
where input $I_i \in \mathcal{D}^m$, $\theta_G^m$ represents the global parameters of site $m$, $\nabla_{\theta_G^m}\left\|\theta_G^m(I_i)\right\|_2^2$ is the gradient of the squared L2 norm of $\theta_G^m(I_i)$. The sensitivity $\Omega_{SS}^m$ shows the extent to which a minor perturbation in $\theta_G^m$ would alter the shape-related segmentation prediction $\theta_G^m(I_i)$, thus during the aggregation of global parameters, it is beneficial to retain parameters with higher $\Omega_{SS}^m$, while updating the parameters with smaller $\Omega_{SS}^m$ by incorporating contributions from other sites as they slightly affect shape knowledge. Therefore, on the server side, we collect $\{\Omega_{SS}^i\}_{i=1}^M$ from all $M$ sites and compute the shape-similarity matrix $\tilde\Omega_{SS}$ by applying the Softmax normalization among $\{\Omega_{SS}^i\}_{i=1}^M$:
\begin{equation}
\label{ssm_server}
    \tilde\Omega_{SS} =\operatorname{Softmax}\{\Omega_{SS}^i\}_{i=1}^M
    = \{\Omega_{SS}^{i'}\}_{i=1}^M
\end{equation}

The global parameters $\theta_G$ for all sites are then updated at the server by:
\begin{equation}
\label{gp_server}
    \theta_G = \theta_G^{all} * \tilde\Omega_{SS}= \sum\nolimits_{i=1}^M\theta_G^i\Omega_{SS}^{i'},
\end{equation}
where $\theta_G^{all}=\{\theta_G^i\}_{i=1}^M$ is the set of global parameters uploaded by all $M$ sites after the local training.

\begin{table*}[tbp]
\centering
\caption{Quantitative results (mean{\tiny$\pm$std}) in different sites (Site-1: EndoVis 2017; Site-2: EndoVis 2018; Site-3: SAR-RARP). $\uparrow$ indicates the higher the score the better, and vice versa. * denotes the personalized federation. Top two results are highlighted in \textbf{bold} and \underline{underlined}.}
\label{tab_quan}
\resizebox{\textwidth}{!}{
\begin{tabular}{p{52pt}|p{28pt}p{28pt}p{28pt}p{28pt}p{1pt}p{32pt}p{32pt}p{32pt}p{32pt}}
\hline
\multirow{2}{*}{Methods} & \multicolumn{4}{c}{Dice(\%) $\uparrow$}    &  & \multicolumn{4}{c}{IoU(\%) $\uparrow$}\\ \cline{2-5}  \cline{7-10} 
                         & \textbf{Average}  & Site-1   & Site-2   & Site-3    &   & \textbf{Average}  & Site-1   & Site-2   & Site-3 \\ \hline
Local Train    & 82.37{\tiny$\pm$0.36}   & 81.45{\tiny$\pm$0.89} & 79.30{\tiny$\pm$0.50} & \underline{86.37}{\tiny$\pm$0.31} & & 73.90{\tiny$\pm$0.21}   & 71.72{\tiny$\pm$0.29} & 71.00{\tiny$\pm$0.69} & \underline{78.97}{\tiny$\pm$0.46}      \\ \hline
FedAvg~\cite{fedavg}   & 83.19{\tiny$\pm$0.13}   & 81.52{\tiny$\pm$0.32} & 82.67{\tiny$\pm$0.32} & 85.38{\tiny$\pm$0.35} & & 74.53{\tiny$\pm$0.15}   & 70.80{\tiny$\pm$0.43} & 75.07{\tiny$\pm$0.32} & 77.71{\tiny$\pm$0.31}    \\
FedSeg~\cite{miao2023fedseg} & 83.18{\tiny$\pm$0.28}   & 80.74{\tiny$\pm$0.71} & \textbf{83.45}{\tiny$\pm$1.02} & 85.34{\tiny$\pm$0.07} & & 74.46{\tiny$\pm$0.37}   & 69.92{\tiny$\pm$1.20} & \textbf{75.83}{\tiny$\pm$1.23} & 77.65{\tiny$\pm$0.05}   \\ 
FedRep*~\cite{fedrep}     & 83.44{\tiny$\pm$0.17}   & 81.90{\tiny$\pm$0.20} & 82.18{\tiny$\pm$1.10} & 86.24{\tiny$\pm$0.67} & & 74.99{\tiny$\pm$0.19}   & 71.69{\tiny$\pm$0.45} & 74.43{\tiny$\pm$1.17} & 78.84{\tiny$\pm$0.80}  \\
pFedLA*~\cite{layerwisedma2022} &82.90{\tiny$\pm$0.46}    &81.24{\tiny$\pm$1.37}   &82.16{\tiny$\pm$1.64}   &85.29{\tiny$\pm$0.28}   & &74.25{\tiny$\pm$0.72}   &71.02{\tiny$\pm$1.96}   &73.77{\tiny$\pm$1.12}   &77.95{\tiny$\pm$0.34}   \\
FedDP*~\cite{feddpwang2023}    & \underline{83.67}{\tiny$\pm$0.13}   & \underline{82.15}{\tiny$\pm$1.13} & 82.84{\tiny$\pm$0.11} & 86.03{\tiny$\pm$0.88} & & \underline{75.22}{\tiny$\pm$0.28}   & \underline{71.96}{\tiny$\pm$1.81} & 75.12{\tiny$\pm$0.10} & 78.59{\tiny$\pm$1.04} \\ \hline
PFedSIS (Ours) & \textbf{85.18}{\tiny$\pm$0.25}   & \textbf{85.08}{\tiny$\pm$0.19} & \underline{83.10}{\tiny$\pm$0.23} & \textbf{87.37}{\tiny$\pm$0.86} & & \textbf{77.33}{\tiny$\pm$0.37}   & \textbf{76.16}{\tiny$\pm$0.18} & \underline{75.56}{\tiny$\pm$0.22} & \textbf{80.26}{\tiny$\pm$1.14}  \\ \hline
\end{tabular}
}
\resizebox{\textwidth}{!}{
\begin{tabular}{p{52pt}|p{28pt}p{28pt}p{28pt}p{28pt}p{1pt}p{32pt}p{32pt}p{32pt}p{32pt}}
\hline
\multirow{2}{*}{Methods} & \multicolumn{4}{c}{ASSD $\downarrow$}    &  & \multicolumn{4}{c}{HD95 $\downarrow$} \\ \cline{2-5}  \cline{7-10}
                         & \textbf{Average}  & Site-1   & Site-2   & Site-3    &   & \textbf{Average}  & Site-1   & Site-2   & Site-3 \\ \hline
Local Train   & 31.61{\tiny$\pm$0.13}   & 22.42{\tiny$\pm$0.18} & 37.73{\tiny$\pm$1.31} & 34.67{\tiny$\pm$1.44}  & & 129.04{\tiny$\pm$5.61}   & 114.28{\tiny$\pm$11.17} & 139.94{\tiny$\pm$6.98} & \underline{132.91}{\tiny$\pm$2.35}      \\ \hline
FedAvg~\cite{fedavg}  & 29.57{\tiny$\pm$1.17}   & 22.36{\tiny$\pm$1.82} & 31.78{\tiny$\pm$3.50} & 34.55{\tiny$\pm$2.60}  & & 122.20{\tiny$\pm$3.05}   & 116.22{\tiny$\pm$7.24} & 109.22{\tiny$\pm$5.21} & 141.18{\tiny$\pm$9.18}      \\
FedSeg~\cite{miao2023fedseg}    & 29.39{\tiny$\pm$1.26}   & 23.59{\tiny$\pm$1.63} & 29.62{\tiny$\pm$1.83} & 34.96{\tiny$\pm$1.26}  & & 124.57{\tiny$\pm$6.21}   & 118.48{\tiny$\pm$6.41} & 108.34{\tiny$\pm$5.49} & 146.87{\tiny$\pm$8.64}   \\ 
FedRep*~\cite{fedrep}  & 28.21{\tiny$\pm$0.41}   & 23.16{\tiny$\pm$0.38} & \underline{28.92}{\tiny$\pm$3.93} & 32.54{\tiny$\pm$3.36}  & & \underline{119.36}{\tiny$\pm$0.75}   & 115.53{\tiny$\pm$3.38} & \underline{104.77}{\tiny$\pm$13.00} & 137.78{\tiny$\pm$12.64}     \\
pFedLA*~\cite{layerwisedma2022} &29.75{\tiny$\pm$1.66}    &22.51{\tiny$\pm$1.72}   &31.08{\tiny$\pm$3.80}   &35.67{\tiny$\pm$0.98}   & &122.12{\tiny$\pm$3.22}   &\underline{112.04}{\tiny$\pm$6.65}   &114.79{\tiny$\pm$8.54}   &139.52{\tiny$\pm$3.86}  \\
FedDP*~\cite{feddpwang2023}  & \underline{27.96}{\tiny$\pm$1.20}   & \underline{22.12}{\tiny$\pm$1.53} & 29.32{\tiny$\pm$0.31} & \underline{32.44}{\tiny$\pm$3.26}  & & 119.40{\tiny$\pm$5.58}   & 112.47{\tiny$\pm$1.89} & 109.93{\tiny$\pm$3.92} & 135.79{\tiny$\pm$12.76} \\ \hline
PFedSIS (Ours)  & \textbf{25.17}{\tiny$\pm$0.69}   & \textbf{18.81}{\tiny$\pm$0.40} & \textbf{27.12}{\tiny$\pm$0.59} & \textbf{29.57}{\tiny$\pm$0.49}  & & \textbf{103.81}{\tiny$\pm$4.44}   & \textbf{89.34}{\tiny$\pm$5.60} & \textbf{100.88}{\tiny$\pm$2.73} & \textbf{121.21}{\tiny$\pm$6.60}   \\ \hline
\end{tabular}
}
\end{table*}

\section{Experiments and Results}
\subsection{Datasets and Evaluation Metrics}
\subsubsection{Datasets}
We use three open-source instrument segmentation datasets of robotic-assisted surgery, to benchmark PFedSIS and evaluate its performance against state-of-the-art FL algorithms. We include EndoVis 2017~\cite{2017allan2019}, EndoVis 2018~\cite{2018allan2020}, and SAR-RARP datasets~\cite{rarp} as three different sites.
\begin{itemize}
    \item \textbf{EndoVis 2017} dataset~\cite{2017allan2019} consists of 10 robotic surgical videos of abdominal porcine procedures with the size of 1280$\times$1024 that are captured by da Vinci Xi systems. We take the first 8 videos (1800 samples) as Site-1's training set and the remaining 2 videos (600 frames) for testing
    \item \textbf{EndoVis 2018} dataset~\cite{2018allan2020} contains 19 sequences of porcine training procedures with the size of 1280$\times$1024 collected by da Vinci X/Xi systems, which are divided into 15 training videos and 4 test videos. We obtain 15 videos with 2235 frames for Site-2's training and 4 videos with 997 frames for testing. 
    \item \textbf{SAR-RARP} dataset~\cite{rarp} is a large-scale dataset containing 50 videos (1920$\times$1080) of suturing the dorsal vascular complex from robotic prostatectomy cases with the da Vinci Si. To prevent this site from dominating in FL development, we selected 22 videos, 17 (4710 frames) as Site-3's training set and 5 (1138 frames) as a test set.
\end{itemize}

\subsubsection{Metrics} 
We utilize four standard and widely used metrics to evaluate model performance: two region-based metrics, Dice and intersection-over-union (IoU), and two distance-based metrics, average symmetric surface distance (ASSD) and 95\% Hausdorff distance (HD95). Higher Dice/IoU and lower HD95/ASSD represent better segmentation results. All metrics are calculated at the original resolution.

\subsection{Implementation Details}
All experiments are implemented in PyTorch on a Tesla V100 GPU. Taking into account the trade-off between efficiency and convergence, we empirically set the maximum communication round $T$ to 200 and local training iterations to 100 during each round, with a batch size of 8. The AdamW optimizer is adopted with an initial learning rate of 5e-4. Input images are resized to 640$\times$512. We evaluate our methods on instrument part segmentation, which includes masks for the instruments' shaft, wrist, and jaws. We report the mean and standard deviation (STD) of all metrics, from three runs with different random seeds.

\subsection{Comparison with State-of-the-Art}
We compare PFedSIS with cutting-edge FL and PFL methods, FedAvg~\cite{fedavg}, FedSeg~\cite{miao2023fedseg}, FedRep~\cite{fedrep}, pFedLA~\cite{layerwisedma2022}, and FedDP~\cite{feddpwang2023}, implemented from released code and original literature, and fine-tuned to the same PVTv2 encoder and FPN decoder for fair comparison. We also use ``Local Train'' where each site trains only on its own dataset without federation.

\begin{figure*}[!thb]
    \centering
    \subfigure[Site-1: EndoVis 2017]{
        \includegraphics[width=\textwidth]{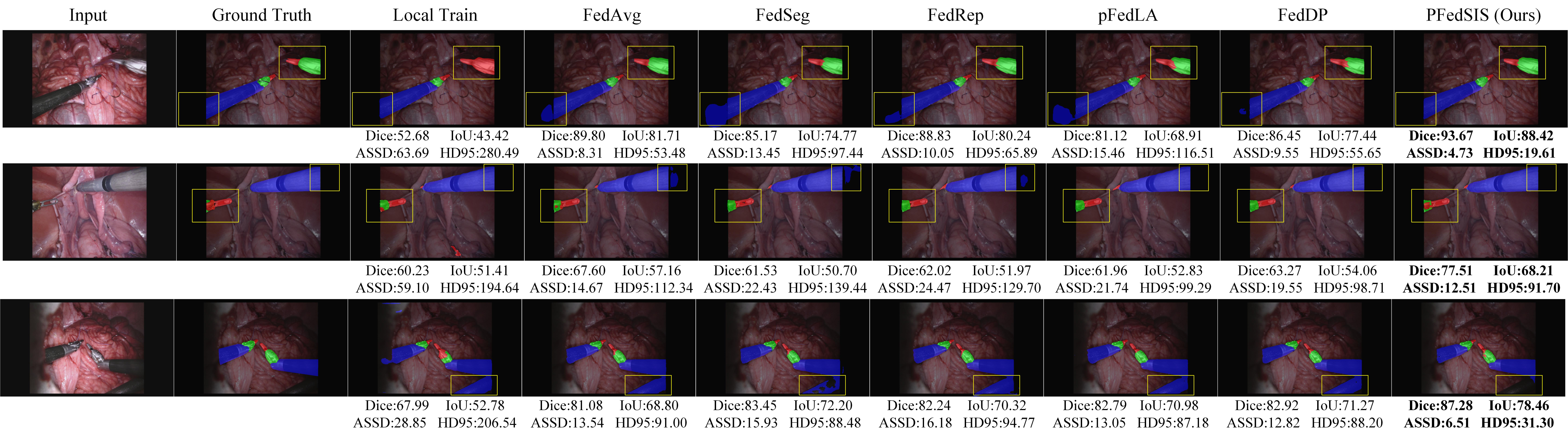}
        \label{fig:sub1}
    }
    \hfill
    \subfigure[Site-2: EndoVis 2018]{
        \includegraphics[width=\textwidth]{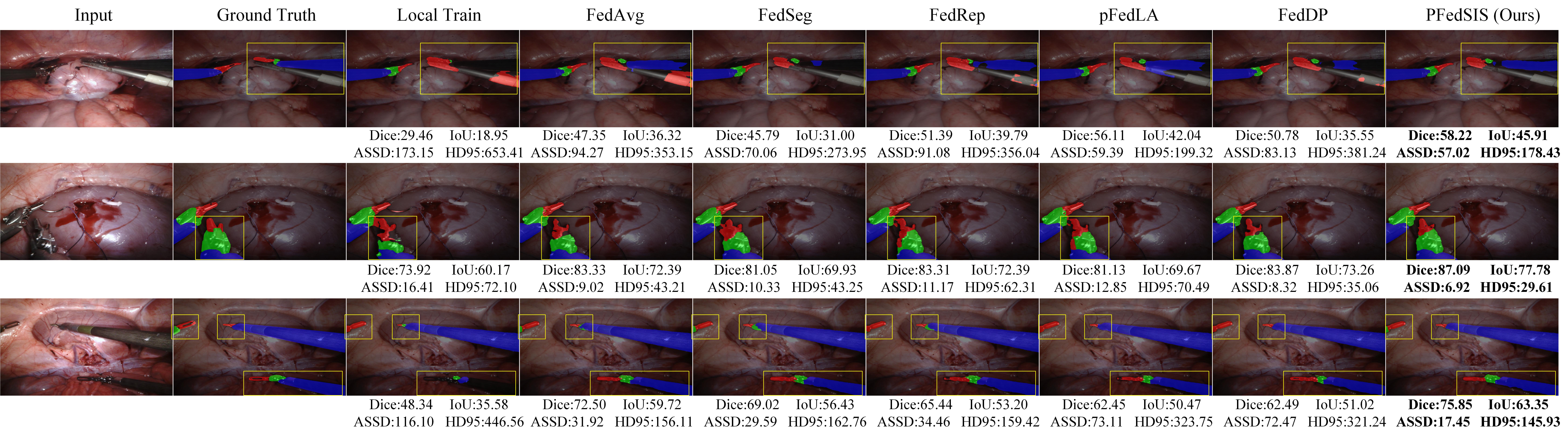}
        \label{fig:sub2}
    }
    \hfill
    \subfigure[Site-3: SAR-RARP]{
        \includegraphics[width=\textwidth]{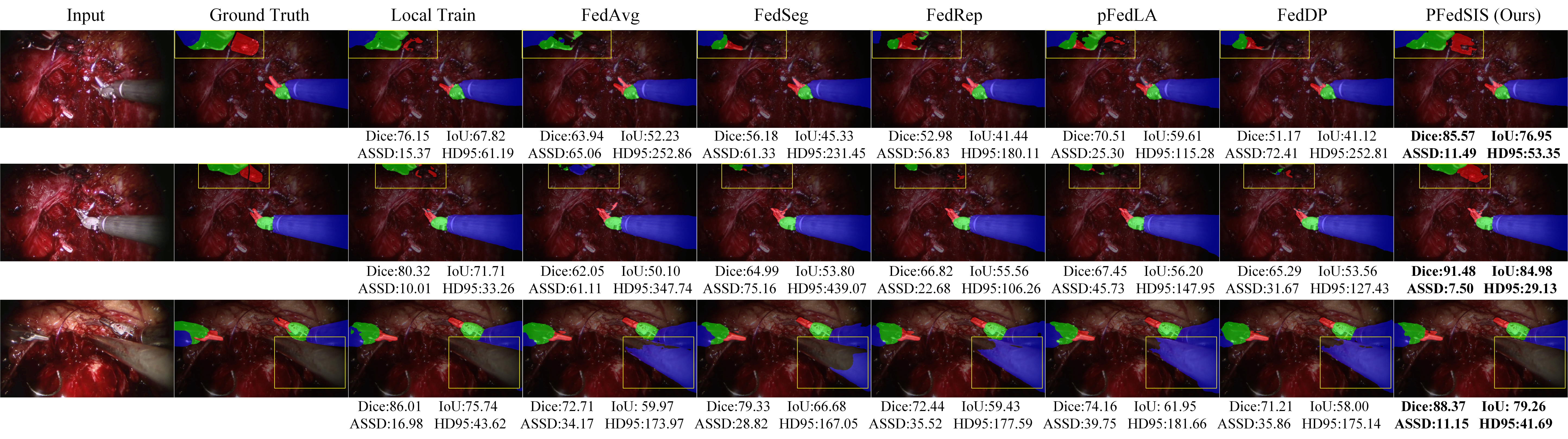}
        \label{fig:sub3}
    }
    \caption{Visual comparison of different methods on three datasets. Yellow boxes highlight the regions where significant differences exist between the methods.}
    \label{fig_visual}
\end{figure*}

\subsubsection{Quantitative Results} 
Table~\ref{tab_quan} lists results on all three sites. Our PFedSIS consistently achieves superior average performance on all metrics, with an increase of 1.51\% in Dice score and 2.11\% in IoU, along with a reduction of 2.79 in ASSD and 15.55 in HD95, compared to suboptimal methods. We prove that the observed improvements are statistically significant and not a result of chance variations by conducting paired t-tests between PFedSIS and the second-best method for averaged Dice, IoU, ASSD, and HD95, yielding p-values of $1.52\times10^{-3}$, $1.07\times10^{-3}$, $8.05\times10^{-3}$, and $9.39\times10^{-3}$, respectively. In all cases p $<5\times10^{-2}$, confirming that the improvements produced by the proposed PFedSIS are statistically significant. The model trained only on local data (``Local Train'') yields good performance, benefiting from the strong Transformer baseline. Nonetheless, all FL methods surpass Local Train in average performance on all metrics, while Site-2 observes large improvements thanks to FL. Interestingly, other FL methods yield inferior Dice, IoU, and HD95 than Local Train on Site-3. This is because cross-site model communication and server fusion impede model learning due to the significant appearance difference in Site-3 data compared to other sites. Conversely, PFedSIS successfully overcomes these challenges, achieving top performance in the in-vivo dataset of Site-3 across all metrics, further affirming its effectiveness.

\subsubsection{Qualitative Results} 
Fig.~\ref{fig_visual} shows typical SIS results. Our PFedSIS yields superior visualization results compared to other methods.\footnote{Additional SIS results are provided in the supplementary Video-S1.} For example, benefiting from personalizing each site's appearance distribution by APE, PFedSIS precisely segments the black borders (left yellow box, 1\textsuperscript{st} row and right yellow box, 2\textsuperscript{nd} row of Fig.~\ref{fig:sub1}) and background objects (3\textsuperscript{rd} row of Fig.~\ref{fig:sub1}) at Site-1, as well as background objects at Site-3 (3\textsuperscript{rd} row of Fig.~\ref{fig:sub3}) as background. Other methods fail to distinguish multiple sites' features from the local site, often misclassifying black borders and background objects as shafts. Additionally, thanks to the mutual shape information of instruments shared by SGE, PFedSIS excels in maintaining shape integrity (2\textsuperscript{nd} row of Fig.~\ref{fig:sub2}; 1\textsuperscript{st} and 2\textsuperscript{nd} rows of Fig.~\ref{fig:sub3}) and enhancing inter-class separation (left yellow box, 2\textsuperscript{nd} row of Fig.~\ref{fig:sub1}; 1\textsuperscript{st} and 3\textsuperscript{rd} rows of Fig.~\ref{fig:sub2}).

\begin{table}[!tbh]
    \centering
    \caption{Computation efficiency analysis of PFedSIS and other federated learning methods. K = $\times 10^3$. w/o means without.}
    \label{tab_computation}
    \resizebox{\columnwidth}{!}{
    \begin{tabular}{c|c|c|c}
        \hline
         \multirow{2}{*}{Methods} &Inference time   &Model Parameters & Train Time  \\
                         &   (milliseconds/frame)   &(Local, Server) & (seconds/round) \\      \hline
         FedAvg &15.20   &(6791.27K, 0)   &106     \\
         FedSeg &15.20   &(6841.062K, 0)    &383     \\
         FedRep &15.20   &(6791.27K, 0)  &106   \\
         pFedLA &15.20   &(6791.27K, 36.6K)   &116\\
         FedDP  &15.20   &(6791.27K, 0) &118   \\ 
         PFedSIS w/o SGE (Ours) &15.20   &(6791.657K, 34.5K) &109    \\
         PFedSIS (Ours)   &15.20  &(6791.657K, 34.5K)   &194   \\
         \hline
    \end{tabular}
    }
\end{table}
\begin{figure}[!tbh]
\centering
\includegraphics[width=0.5\columnwidth]{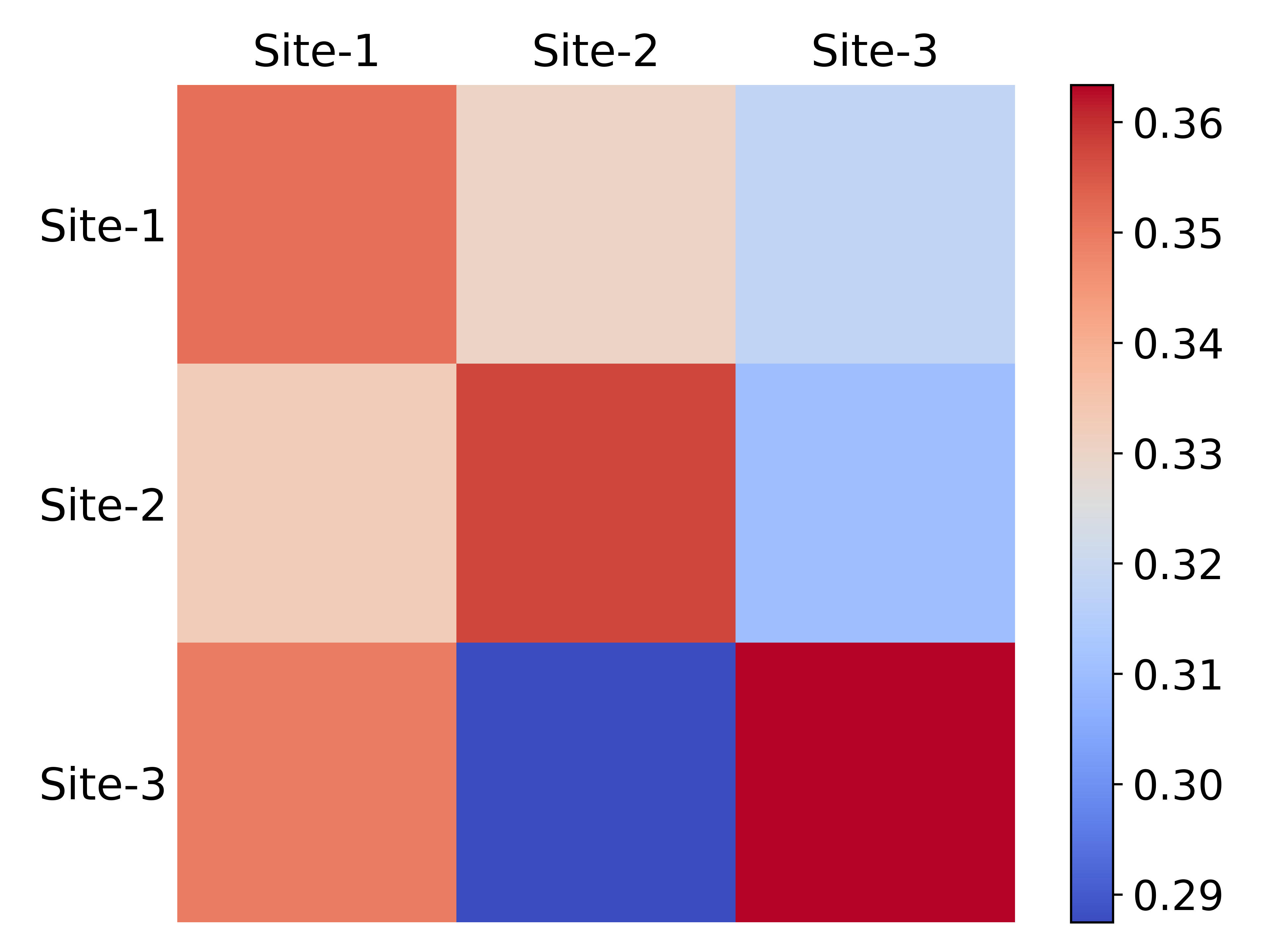}
\caption{The heatmap of aggregation matrices $\Omega_{HN}^1, \Omega_{HN}^2, \Omega_{HN}^3$ of the regulation head $H_{ar}$ generated by all three sites' hypernetworks. X-axis and Y-axis show the IDs of sites. Each row represents the aggregation weights of that site's regulation head.}
\label{fig_heatmap}
\end{figure}

\subsubsection{Efficiency Analysis}
Table~\ref{tab_computation} shows the computation complexity of all methods. Notably, the appearance head $H_{ar}$ (with 0.387K parameters) and hypernetwork (with 34.5K parameters) are only used in the training process, so PFedSIS does not introduce additional processing burden during inference, maintaining a swift inference time of 15.20 milliseconds per frame—equivalent to other methods, and enabling real-time ($\sim$66 frames per second) deployment. Although PFedSIS requires additional training time due to the computation of sensitivity in SGE, it achieves significant performance improvement compared to other methods. Besides, our ``PFedSIS w/o SGE'' variant has comparable training time with FedAvg, yet outperforms it by a considerable margin (+1.67\% Dice, +2.2\% IoU, -4.12 ASSD, and -17.14 HD95, see 1\textsuperscript{st} and 4\textsuperscript{th} rows of Table~\ref{tab_ab}). In summary, the proposed method achieves a favorable balance between computational complexity and performance compared to other approaches.

\subsection{Ablation Study}
\subsubsection{Effectiveness of Key Components in PFedSIS} 
Ablation results on PFedSIS components are listed in Table~\ref{tab_ab}. Each proposed component enhances model performance and complements the others. Specifically, GPD brings +1.19\% IoU and -11.81 HD95 performance gains. Introducing SGE and APE to GPD further gains +0.96\% IoU, -5.56 HD95 and +1.01\% IoU, -5.33 HD95 respectively. After removing all proposed modules (GPD, SGE, APE), PFedSIS becomes FedAvg, where all parameters are merged and averaged through the server. Fig.~\ref{fig_heatmap} visualizes the $\Omega_{HN}$ heatmap of the appearance regulation head generated by different sites' hypernetworks. It can be observed that Site-1 and Site-2, which have smaller appearance discrepancies compared to Site-3, show a greater tendency to merge each other's personalized parameters. Additionally, each site's aggregation matrix has distinct weights compared to others. These observations underscore the ability of APE to learn appearance discrepancies across different sites.

\begin{table}[!tbp]
    \centering
    \caption{Ablation results for key components in PFedSIS, i.e. GPD, APE, and SGE. Results are averaged across sites.}
    \label{tab_ab}
    \resizebox{\columnwidth}{!}{%
    \begin{tabular}{ccc|cccc}
        \hline
        GPD    &SGE &APE &Dice(\%) $\uparrow$   &IoU(\%) $\uparrow$   &ASSD $\downarrow$   &HD95 $\downarrow$   \\ \hline
        &   &   &83.19   &74.53   &29.57   &122.20 \\
        \checkmark   &    &     &83.99   &75.72    &26.41   &110.39 \\ 
        \checkmark &\checkmark   & &84.67   &76.68  &25.54   &\underline{104.83} \\           
        \checkmark &   &\checkmark   &\underline{84.86}   &\underline{76.73}  &\underline{25.45}   &105.06\\
        \checkmark &\checkmark   &\checkmark   &\textbf{85.18}   &\textbf{77.33}    &\textbf{25.17}   &\textbf{103.81} \\
        \hline
        \end{tabular}
        }
\end{table}

\subsubsection{Impact of Different Personalized Settings in GPD} 
We progressively personalize the model along the channel dimension of the FPN decoder and the dimension of the query, key, and value heads in the MSA to explore the efficacy of GPD. As shown in Table~\ref{tab_ab2},  decoupling the decoder parameters yields an increase of 0.26\% in IoU and a reduction of 1.81 in HD95, compared to the baseline FedAvg method (1\textsuperscript{st} row). Further personalizing the query, key, and value embedding heads leads to additional +0.93\% IoU and -10 HD95 performance improvement, compared to only decoupling the decoder.

\begin{table}[!tbp]
    \centering
    \caption{Ablation results for different GPD settings. Check marks refer to personalizing half channels in decoder and half q, k, v heads in MSA. Results are averaged across sites.}
    \label{tab_ab2}
    \resizebox{\columnwidth}{!}{
        \begin{tabular}{cccc|cccc}
        \hline
        Decoder &q   &k  &v  &Dice(\%) $\uparrow$   &IoU(\%) $\uparrow$   &ASSD $\downarrow$   &HD95 $\downarrow$   \\ \hline
           &   &   &   &83.19   &74.53   &29.57   &122.20   \\
        \checkmark  &   &   &   &83.30   &74.79   &29.21   &120.39   \\
        \checkmark  &\checkmark   & &   &83.41   &75.12   &\underline{27.00}   &\underline{116.07}  \\
        \checkmark  &   &\checkmark   &   &\underline{83.53}   &\underline{75.38}   &27.33 &118.33   \\
        \checkmark  &   &   &\checkmark   &83.35   &74.88   &28.35  &119.09   \\
        \checkmark  &\checkmark   &\checkmark   &\checkmark   &\textbf{83.99}   &\textbf{75.72}   &\textbf{26.41}   &\textbf{110.39}    \\
        \hline
        \end{tabular}
        }
\end{table}

\section{Discussion}
Surgical instrument segmentation is essential for robotics-assisted surgery and computer-aided interventions. It can enhance surgical precision, reduce the risk of accidental damage to vital structures and elevate understanding of operating field. Developing an expert segmentation model necessitates a substantial amount of annotated data and relying exclusively on data from a single clinical site is inadequate to meet training requirements, while the bias inherent in single-centre data impairs the model's generalization ability. Moreover, ethical and privacy policies must be satisfied with controlled data access.

This paper represents the initial effort to use PFL for surgical instrument segmentation, which enables collaborative training without compromising patient privacy while ensuring site personalization through the aggregation of knowledge from various sites. Unlike other PFL approaches, our PFedSIS proposes a GPD to consider head-wise personalization for the multi-headed self-attention mechanism, allowing for more granular customization of segmentation models. Furthermore, PFedSIS propose APE and SGE based on appearance discrepancy and shape similarity priors to exploit inter-site inconsistency and consistency, respectively, achieving efficient segmentation performance on each site. On the one hand, APE utilizes appearance regulation to guide personalized parameters toward the local site's appearance distribution and build individual hypernetworks for each site to customize personalized parameter updates. On the other hand, SGE applies cross-style shape consistency to the input and its segmentation map, which enables global parameters to focus on mutual shape information across different sites, and preserves global parameters crucial for shape information via shape-similarity update. Comparison experimental results show that the proposed PFedSIS surpasses previous works to a significant extent. 

PFedSIS focuses on the SIS task, which serves as a foundational task supporting numerous downstream applications, such as image-guided automated suturing manipulations \cite{zhao2021one}, tool pose estimation~\cite{sestini2021kinematic}, instrument tracking~\cite{cheng2021deep}, trajectory prediction~\cite{toussaint2021co}, intraoperative navigation~\cite{liu2023instrumentnet}, and contributes to the next generation of automation in operating intelligence~\cite{lu2021toward}. Surgical instruments are non-deformable, articulated rigid bodies, which allows SGE to exploit a common shape prior across sites. Segmentation tasks on deformable-shape surfaces, such as tumor segmentation with distinctive abnormal structures will pose a challenge for extracting mutual shape information. We will extend our method to tissue (with distinct shapes) segmentation tasks and use generative methods (e.g. diffusion models), to introduce shape perturbations. 

\section{Conclusions}
This paper presents PFedSIS, a novel PFL method for SIS, capitalizing on visual trait priors of appearance disparity and instrument shape similarity. It first decouples the embedding heads of MSA and channels of convolution layers into personalized and global via GPD. Then, APE enhances personalized parameters tailored to each site's appearance representation and utilizes inter-site appearance inconsistency via hypernetwork for personalized parameters update, while SGE enables global parameters to maintain shape information via cross-style consistency and effectively share mutual shape information via shape-similarity update. Experimental results on three publicly available datasets show that the proposed PFedSIS achieves superior performance in both quantitative assessment and visual interpretation, while its computation requirements during prediction remain low and comparable to other methods.

\bibliographystyle{IEEEtran}
\bibliography{refs.bib}
\end{document}